\documentclass{article} 
\usepackage{collas2024_conference,times}
\usepackage{easyReview}
\usepackage{subcaption}
\usepackage{algorithm}
\usepackage{algorithmic}
\usepackage{natbib}
\usepackage{booktabs}
\usepackage{multirow}

\usepackage{amsmath,amsfonts,bm}

\def\eqref#1{equation~\ref{#1}}

\def\1{\bm{1}}

\DeclareMathAlphabet{\mathsfit}{\encodingdefault}{\sfdefault}{m}{sl}
\SetMathAlphabet{\mathsfit}{bold}{\encodingdefault}{\sfdefault}{bx}{n}

\usepackage{hyperref}
\hypersetup{
    colorlinks=true,
    linkcolor=red,
    filecolor=magenta,
    urlcolor=blue,
    citecolor=purple,
    pdftitle={Overleaf Example},
    pdfpagemode=FullScreen,
    }

\title{Decoupled Prompt-Adapter Tuning for Continual Activity Recognition}

\author{Di Fu \\
School of Computing\\
National University of Singapore\\
Singapore \\
\texttt{difu@u.nus.edu} \\
\And 
Thanh Vinh Vo \\
School of Computing\\
National University of Singapore\\
Singapore \\
\texttt{votv@comp.nus.edu.sg} \\
\And 
Haozhe Ma \\
School of Computing\\
National University of Singapore\\
Singapore \\
\texttt{haozhe.ma@u.nus.edu} \\
\And 
Tze-Yun Leong \\
School of Computing\\
National University of Singapore\\
Singapore \\
\texttt{leongty@comp.nus.edu.sg}
}

\collasfinalcopy 

\begin{document}

\maketitle

\begin{abstract}
Action recognition technology plays a vital role in enhancing security through surveillance systems, enabling better patient monitoring in healthcare, providing in-depth performance analysis in sports, and facilitating seamless human-AI collaboration in domains such as manufacturing and assistive technologies. The dynamic nature of data in these areas underscores the need for models that can continuously adapt to new video data without losing previously acquired knowledge, highlighting the critical role of advanced continual action recognition. To address these challenges, we propose Decoupled Prompt-Adapter Tuning (DPAT), a novel framework that integrates adapters for capturing spatial-temporal information and learnable prompts for mitigating catastrophic forgetting through a decoupled training strategy. DPAT uniquely balances the generalization benefits of prompt tuning with the plasticity provided by adapters in pretrained vision models, effectively addressing the challenge of maintaining model performance amidst continuous data evolution without necessitating extensive finetuning. DPAT consistently achieves state-of-the-art performance across several challenging action recognition benchmarks, thus demonstrating the effectiveness of our model in the domain of continual action recognition.
\end{abstract}

\section{Introduction}

The widespread deployment of cameras has significantly broadened the scope and influence of action recognition technology across multiple sectors. This technology is essential for boosting safety via security and surveillance, offering vital patient care in healthcare, and providing in-depth performance analyses in sports~\citep{kong2022human}. It also enables robots to quickly perceive and respond to human actions during human-AI collaborations~\citep{akkaladevi2015action}, thereby enhancing the collaboration and efficiency between humans and AI in contexts such as manufacturing and assistive technologies. In this context, the importance of continual learning (CL) emerges, driven by the technological imperative to synchronize with the dynamically evolving nature of human activities and interactions. Defined as the ability of the model to incorporate new information from an ongoing data stream while retaining previously acquired knowledge, continual learning plays an essential role in this arena. It adeptly navigates the complex and heterogeneous spectrum of actions an action recognition algorithm encounters, enabling adaptation to emerging actions and contexts over time without compromising the recognition of previously observed actions. This flexibility is crucial for sustaining the efficacy and relevance of action recognition systems across their extensive applications, thereby establishing continual learning as a foundational element for the advancement and persistent applicability of action recognition technologies.

While continual learning has made significant progress in recent years, the specific challenge of continual action recognition, which involves learning from dynamic video data streams without forgetting previously acquired knowledge, remains a difficult problem. On the one hand, the majority of CL methodologies are primarily devised for static images ~\citep{wang2022learning,wang2022dualprompt,smith2023coda} , rendering them inadequately equipped to confront the unique challenges presented by video data. These challenges encompass the high-dimensional nature of video data, temporal dependencies, and the significant variability across video sequences. Such challenges can either impede a model's ability to adapt to new tasks or induce rapid changes that lead to catastrophic forgetting. 
On the other hand, many existing methods tailored specifically for continual learning in the context of video~\citep{villa2023pivot,pei2022learning} require the retention of data for new classes. These approaches inevitably lead to escalating memory costs, which become prohibitively expensive given the typically large size of video data. Consequently, the evolution of models in this domain is hampered by a delicate balance between adaptability and memory efficiency.

Recent advancements in large pre-trained models~\citep{bao2021beit,dosovitskiy2020image,radford2021learning} have spurred the development of lightweight tuning techniques, such as Adapters~\citep{houlsby2019parameter} and Prompt Tuning~\citep{li2021Prefix, lester2021power}, addressing the cost and time-intensive nature of fine-tuning large models. These methods enable the refinement of pre-trained models for new tasks with minimal increases in parameters, significantly reducing the computational and memory demands typically associated with training models from scratch. Furthermore, it has been observed that by tuning fewer parameters, these methods exhibit reduced susceptibility to forgetting and enhanced generalization capabilities~\citep{vander2023using}, albeit with limited plasticity. This insight has catalyzed the trend of employing pre-trained models in continual learning scenarios~\citep{wang2022dualprompt,wang2022learning}. 
Despite the distinct advantages of adapters and prompt tuning, where adapters mitigate catastrophic forgetting compared to conventional fine-tuning, and prompt tuning improves stability and generalizability, 
their standalone applications exhibit inherent limitations. 
While adapters have shown promise in continual learning, they struggle with rapid task specialization, as they require a certain amount of data to effectively adapt to new tasks. On the other hand, prompt tuning exhibits a slower rate of adaptation to new tasks, as the prompts need to capture task-specific information while maintaining the model's stability. This slow adaptation can lead to homogeneity in the learned representations, limiting the model's ability to distinguish between different tasks~\citep{gao2023unified}. DPAT addresses these limitations by combining the strengths of both adapters and prompt tuning in a decoupled training strategy, allowing for efficient task specialization while preserving the model's stability and generalization capabilities. By integrating adapters, DPAT exhibits enhanced adaptability to current tasks, achieving improved spatial-temporal adaptation to new tasks. Incorporation of learnable prompts further augments the model's stability, making it less susceptible to forgetting. Moreover, we employ an enhanced decoupled training strategy for prompts and adapters, leveraging the strengths of both components. This strategy streamlines the adaptation to new activities and fortifies generalization, while significantly lowers both computational and memory requirement.

\textbf{Contributions:} 
 Our contributions are outlined as follows:
\begin{itemize}
    \item[\textbf{(i)}] We introduce \textbf{\underline{D}}ecouple \textbf{\underline{P}}rompt-\textbf{\underline{A}}dapter \textbf{\underline{T}}uning (DPAT), a framework developed to improve the performance of pretrained image encoders in the context of Continual Action Recognition.
    \item[\textbf{(ii)}] By leveraging the intrinsic capabilities of pre-trained Vision Transformer (ViT) in conjunction with adapters and prompt tuning, we provide a robust solution that ensures the retention of previously acquired knowledge while seamlessly adapting to new spatial-temporal tasks. 
    \item[\textbf{(iii)}] We substantiate that the designed dual-stage training strategy employed for adapters and spatial-temporal prompts plays a crucial role in harmonizing the model's plasticity and generalization capabilities, which safeguard against the adverse effects of new information on previously mastered tasks.
    \item[\textbf{(iv)}] In our extensive experiments across challenging datasets, DPAT consistently achieve state-of-the-art performance, showcasing its superiority in handling the fine-grained action recognition tasks.
\end{itemize} 
\begin{figure}[ht]
  \centering
  \begin{subfigure}[t]{0.57\textwidth}
\includegraphics[width=\textwidth]{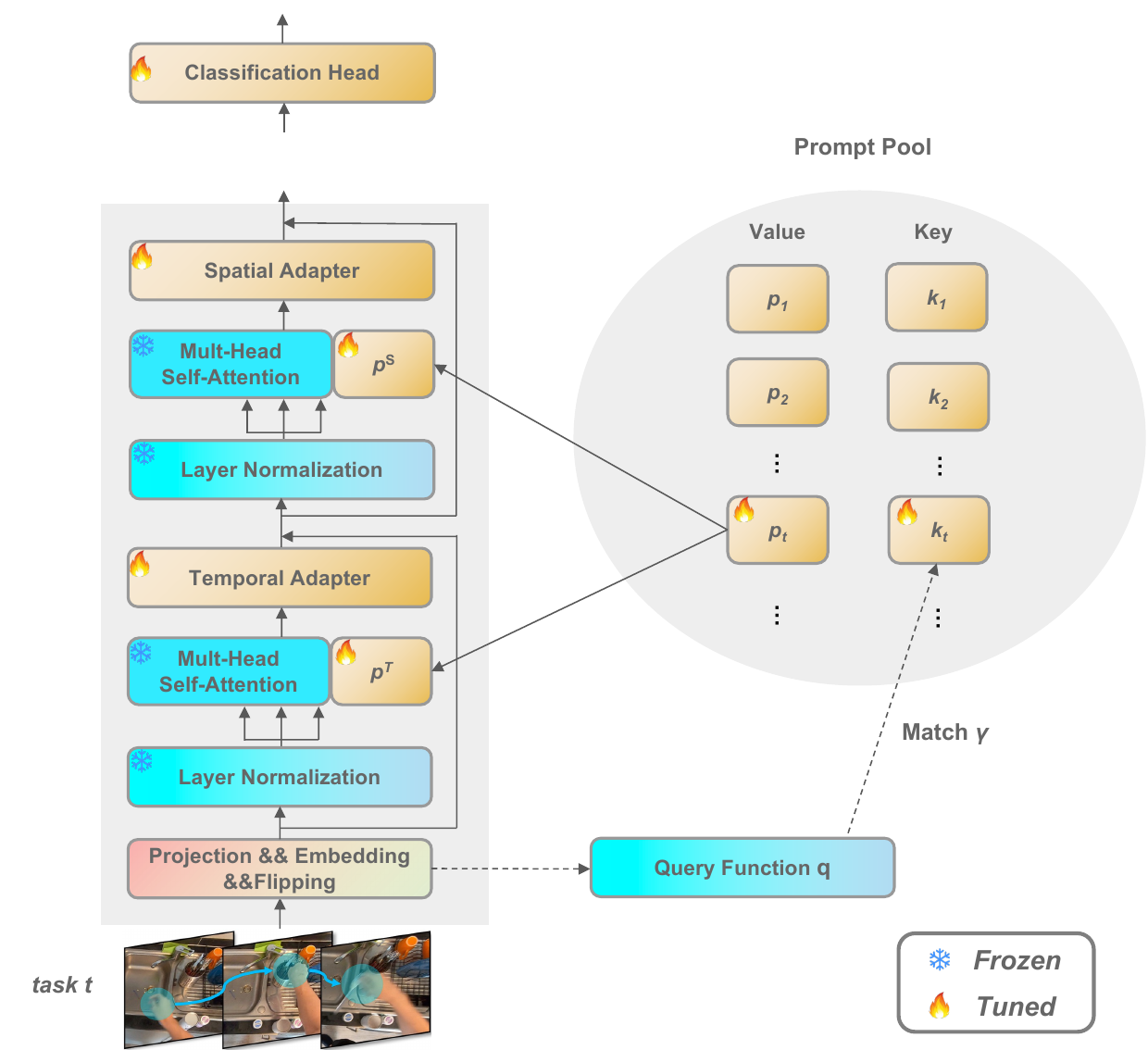}
    \caption{Depiction of the model architecture designed for continual video action recognition, leveraging spatial and temporal adapters in conjunction with two prompts ($\boldsymbol{p}^S$ and $\boldsymbol{p}^T$), all operating within a frozen Multi-Head Self-Attention (MSA) in pre-trained ViT. Meanwhile, the inclusion of a learnable key ($k_t$), utilized in the model's key-query matching mechanism, facilitates optimal prompt selection and mitigates forgetting during inference.}
    \label{fig:sub1}
  \end{subfigure}
  \hfill 
  %
  \begin{subfigure}[t]{0.36\textwidth}
    \includegraphics[width=\textwidth]{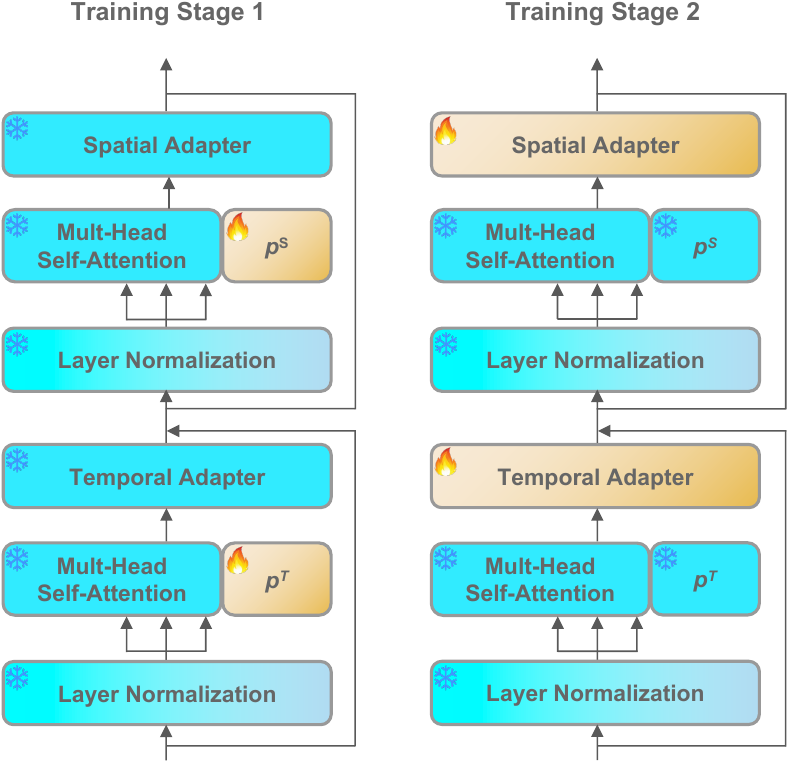}
    \caption{Demonstration of our decoupled training strategy: we firstly freeze the adapter to establish a stable and generalizable foundation through Prefix tuning, then refine task-specific capabilities by focusing on adapter tuning while preserving the initial prompts for balanced adaptation and generalization.}
    \label{fig:sub2}
  \end{subfigure}
  
  \caption{Overview of the proposed Decoupled Prompt-Adapter Tuning (DPAT) approach: (a) Model architecture integrating adapters and prefix prompts to facilitate adaptation to new tasks;  (b) Decoupled training paradigm designed to bolster knowledge preservation through phase-separated optimization of the model components}
  \label{fig:overall}
\end{figure}

\section {Related Work}

\textbf{Continual Learning.} Continual Learning endeavors to mitigate the issue of catastrophic forgetting. Inspired by the hippocampus's replay mechanism in the human brain, memory replay methods adopt either the conservation of real samples for future replay~\citep{rolnick2019experience,rebuffi2017icarl,buzzega2020dark} or the employment of generative models to create samples that emulate previous task distributions~\citep{shin2017continual,ostapenko2019learning}. These methodologies, while effective, are constrained by substantial storage requirements and the complexity inherent in producing high-fidelity synthetic samples, especially when dealing with high-dimensional data such as video content. On the other hand, regularization-based methods~\citep{kirkpatrick2017overcoming,aljundi2018memory,zenke2017continual,li2017learning} present an alternative strategy aimed at preserving essential weight configurations from preceding tasks. This is achieved through an array of analytical instruments, including the Fisher information matrix, gradient assessments, and uncertainty metrics, which are employed to evaluate and rank the significance of weights. These approaches, which obviate the need for the storage of additional data, offer a distinct advantage in terms of efficiency and privacy. However, empirical evidence suggests that excessive regularization can impair the model's ability to generalize effectively across complex tasks. This limitation underscores the delicate balance required in the application of regularization techniques, ensuring that the preservation of prior knowledge does not come at the expense of the model's adaptability and learning capacity for future tasks. 

\textbf{Continual Learning for Large Pre-trained Model.}  Recently, the emergence of large pre-trained models has catalyzed the development of prompt-based approaches~\citep{wang2022dualprompt,wang2022learning,smith2023coda} in the realm of continual learning (CL). Pioneering works such as DualPrompt~\citep{wang2022dualprompt} and L2P~\citep{wang2022learning} leverage a minimal set of trainable prompts instead of directly modifying the model's encoder parameters. These methods align input data with appropriate prompts via a task-agnostic, local clustering-like optimization process by assembling a pool of prompts from which selections are made. On the other hand, in the sphere of prompt-based learning for Vision-Language Models (VLMs), significant strides have been made with contributions from works like ProGrad~\citep{zhu2023prompt}, CoCoOp~\citep{zhou2022conditional}, and CoOp~\citep{zhou2022learning}. These efforts have introduced strategies to mitigate forgetting and enhance the adaptability of prompts for downstream tasks. ProGrad, for example, unveils a progressive prompt training strategy that gradually escalates prompt length. CoCoOp employs a conditional framework for generating input-dependent prompts, while CoOp embraces a cooperative learning methodology to simultaneously optimize image and text prompts. These techniques provide valuable perspectives on employing prompt learning to boost VLMs' performance within continual learning scenarios.   However, it is worth noting that the application of these prompt tuning methods has predominantly been explored within the context of static image learning. The extension of these methods to video processing remains an area ripe for exploration.

\textbf{Continual Learning for Activity Recognition.}  
Several studies have addressed the challenges of temporal dynamics and complexity in video data to improve continual learning frameworks' knowledge retention across video tasks. \citet{park2021class} leveraged time-channel information for weighted knowledge distillation, aiming to better encode temporal dynamics and combat forgetting. \citet{villa2023pivot} introduced the PIVOT model, which enhances temporal modeling and mitigates forgetting through the integration of spatial prompts, memory replay, and a transformer encoder, significantly boosting video classification accuracy. Meanwhile, \citet{pei2022learning} developed a memory-efficient approach by creating a condensed frame representation for each representative video from previously seen classes. While these methods demonstrate innovative techniques to mitigate forgetting and manage memory, they share a reliance on storing samples, indicating that their strategies are not rehearsal-free. Conversely, ST-Prompt~\citep{pei2023space} presented a novel rehearsal-free approach in video continual learning by leveraging pre-trained vision-language models with temporal prompts for temporal information encoding. While this approach, necessitating additional inputs from a text encoder in CLIP~\citep{radford2021learning} for distribution alignment, may struggle to capture videos with intricate temporal dynamics solely through prompts, our methods circumvents the need for a text encoder and instead employs adapters to bolster temporal modeling capability
\section{Preliminary}


\subsection{Continual Action Recognition}

In the domain of action recognition, continual learning frameworks are tailored to incrementally adapt to a series of data streams, denoted as \(\{D_1, D_2, \ldots, D_T\}\). Each stream \(D_t\) at stage \(t\) comprises \(N_t\) labeled video clips \(\{(v_t^b, y_t^b)\}_{b=1}^{N_t}\), where \(v_t^b\), a video clip, is characterized within \(\mathbb{R}^{T \times H \times W \times C}\), and \(y_t^b\) represents the label of the video clip \(v_t^b\). In this context, \(v_t^b\) is defined in a four-dimensional space, with \(T\) representing the number of frames, \(H\) the height, \(W\) the width, and \(C\) the number of channels. The label \(y_t^b\) is a categorical variable that identifies the class of the video clip \(v_t^b\). In the class incremental learning framework, distinct, non-overlapping class groups are introduced at each stage (\(Y_t\)), with the constraint that \(Y_i \cap Y_j = \emptyset\) for any \(i \neq j\). This setup mandates that the model, \(f_\theta\), not only assimilates new data from the current dataset \(D_t\) while retaining the ability to accurately classify across an expanding set of classes, encapsulated as \(\tilde{Y}_t = \bigcup_{i=1}^{t} Y_i\). The primary objective is to optimize a singular model's average classification accuracy across all tasks.

\subsection{Continual Learning with Prefix-Tuning}
In the realm of continual learning, Prefix-tuning~\citep{li2021Prefix} has become a pivotal strategy for adapting Transformer models to new tasks with minimal retraining effort. Let the input to the Multi-Head Self-Attention (MSA) layer be \(\boldsymbol{h} \in \mathbb{R}^{L \times D}\), and we further denote the input query, key, and values for the MSA layer to be \(\boldsymbol{h}_Q\), \(\boldsymbol{h}_K\), and \(\boldsymbol{h}_V\), respectively. The prompt parameter \(\boldsymbol{p} \in \mathbb{R}^{L_p \times D}\) is divided into \(\{\boldsymbol{p}^K, \boldsymbol{p}^V\} \in \mathbb{R}^{\frac{L_p}{2} \times D}\) and prepended to \(\boldsymbol{h}_K\), and \(\boldsymbol{h}_V\) in the MSA as:
\begin{equation}
\label{eqn:Pre-T}
f_{\text {Pre-T }}(\boldsymbol{p}, \boldsymbol{h})=\operatorname{MSA}\left(\boldsymbol{h}_Q,\left[\boldsymbol{p}^k ; \boldsymbol{h}_K\right],\left[\boldsymbol{p}^v ; \boldsymbol{h}_V\right]\right).
\end{equation}
To address catastrophic forgetting during Prefix tuning, state-of-the-art methods such as L2P~\citep{wang2022learning} and DualPrompt~\citep{wang2022dualprompt} utilize a key-value pair query strategy for dynamically selecting instance-specific prompts from a pool. Each prompt \(p_m\) is linked to a learnable key \(k_m\), with \(M\) denoting the pool's size, selected based on cosine similarity against an input-conditioned query \(q(x)\), thereby identifying the key \(k_m\) with the highest similarity \(\gamma(q(x), k_m)\). During the test, the prompt embedding task-specific information is selected by \(\arg\min_m \gamma(q(x), k_m)\), ensuring that the keys, embedded with task-specific knowledge during training, are precisely matched to the input for inference. However, these approaches were primarily tested for dealing with static images and do not accommodate the temporal information crucial for video action recognition, rendering them inadequate for such applications, whereas our proposed model seeks to address such limitations by incorporating additional adapters.

\section{Method}
Our proposed approach is illustrated in Figure~\ref{fig:overall}. We begin by explaining the configuration of the adapter and prompt in our model in Section~\ref{sec:adapter_prompt_configuration}, where we discuss how these components work with the MSA layer in the pretrained image encoder. In Section~\ref{sec:decoupled_training_process}, we describe our decoupled training process. In Section~\ref{sec:redesigned_query_key_matching}, we outline the redesigned query-key matching loss. Finally, we detail the training objectives in Section~\ref{sec:training_objectives}.

\subsection{Position of Adapter and Prompt}\label{sec:adapter_prompt_configuration}
Figure~\ref{fig:sub1} illustrates our approach, diverging from recent methods that append an additional temporal model atop the ViT backbone to capture temporal dynamics. Instead, we deploy adapters to refine the model's inherent processing capabilities. Specifically, our model incorporates a spatial adapter, \textit{Adapter-S}, with a spatial prompt \(\boldsymbol{p}^S\), and a temporal adapter, \textit{Adapter-T}, with a temporal prompt \(\boldsymbol{p}^T\). This configuration maintains the original functionality of the MSA layer within the pre-trained image encoder, leveraging its inherent strengths while minimizing modifications. Expanding upon \citet{villa2023pivot}'s demonstration of repurposing pre-trained image encoders for temporal analysis, our approach distinctively augments the pre-trained image model with adapters and prompts. This not only facilitates exhaustive extraction of spatial and temporal information with adapter but also leverages prompts to mitigate model forgetting and enhance stability. Following the dual-prompt approach, \(\boldsymbol{p}^T\) includes a task-agnostic prompt, \(\boldsymbol{g}^T\), positioned at a shallower layer to capture task-independent information, which is learned and shared across all tasks, and a task-specific prompt, \(\boldsymbol{e}^T)\), aimed at collecting task-related information. Similarly, \(\boldsymbol{p}^S\) is structured with its task-agnostic component, \(\boldsymbol{g}^S\), and task-specific component, \(\boldsymbol{e}_S\). Furthermore, for any task \(t\), the task-specific prompt \(\boldsymbol{e}_t = \{\boldsymbol{e}_t^T, \boldsymbol{e}_t^S\}\) is associated with a task-specific key, \(\boldsymbol{k}_t\), a learnable parameter designed to capture the distinctive features of task \(t\). During inference, a pre-defined query function \(q\) is employed to perform key-query matching to facilitate the selection of the appropriate task-specific prompt to be prepended to the model. 

Specifically, upon receiving a video patch embedding, $\boldsymbol{z} \in \mathbb{R}^{T \times (N+1) \times D}$, it is first reshaped into $\boldsymbol{z}^T \in \mathbb{R}^{(N+1) \times T \times D}$, where $T$ represents the temporal dimension, or the number of frames. This reshaped embedding, $\boldsymbol{z}^T$, is subsequently fed into a pre-trained MSA layer, enabling it to learn the complex relationships among the $T$ frames. Concurrently, we employ Prefix tuning by prepending a prompt parameter $\boldsymbol{p}^T$ to the MSA layer. Following the extraction of temporal features, the output from the temporal adapter is reshaped back to \(\mathbb{R}^{T \times (N+1) \times D}\) to enable the extraction of spatial features. This reshaped output is then processed by the spatial adapter, which is specifically designed to enhance the model's capabilities in spatial analysis. Similar to the approach taken with the temporal domain, we prepend a learnable prompt \(\boldsymbol{p}^S\) to the spatial adapter's output. Given the aforementioned model structure, the forward process in our proposed model can be written as:
\begin{equation}
\label{eqn:forward}
\begin{aligned}
h^l_T &= h^{l-1} + \textit{Adapter-T}\left(f_{\text{Pre-T}}(\boldsymbol{p}^T, \text{LN}(h^{l-1}))\right), \\
h^l_S &= h^l_T + \textit{Adapter-S}\left(f_{\text{Pre-T}}(\boldsymbol{p}^S, \text{LN}(h^l_T))\right),
\end{aligned}
\end{equation}
where \(h^l_T\) and \(h^l_S\) denote the output after temporal and spatial feature extraction, respectively. \(LN\) is the layer normalization. The function \(f_{\text{Pre-T}}\) is defined as in Equation~\ref{eqn:Pre-T}.

\subsection{Decoupled Prompt-Adapter Tuning}\label{sec:decoupled_training_process}
Despite the theoretical advantage of adapters being less susceptible to catastrophic forgetting compared to traditional fine-tuning, their standalone application does not completely circumvent the challenges associated with rapid task specialization. Conversely, employing Prefix tuning directly, although it enhances model stability and generalizability, it has been observed that Prefix tuning exhibits a slower adaptation rate to new tasks and is prone to homogeneity~\citep{gao2023unified}, thereby constraining the adaptability required for varied and intricate tasks. These observations form the core motivation for our proposed strategy, Decoupled Prefix Prompt and Adapter Tuning, which aims to harness the complementary strengths of both adapter and Prefix tuning in a unified framework.

Figure~\ref{fig:sub2} illustrates the proposed decoupled training strategy, the training process is delineated into two distinct phases, meticulously designed to balance adaptability with generalizability:

\textbf{First Stage: Prefix Tuning.} Initially, Prefix tuning is employed to provide the model with a stable and generalizable base. Learnable prompts encapsulate the task-specific information, reducing the immediate adaptation pressure for adapter and providing the model with a robust understanding of the task. This stage is crucial for setting the stage for specialized adaptation.

\textbf{Second Stage: Adapter Tuning.} Subsequently, the focus shifts to adapter tuning, emphasizing task-specific refinement and adaptation. By maintaining the prompt learned in the initial phase, we preserve the generalization and stability benefits of Prefix tuning, while the adapter module targets task-specific learning. This approach aims to strike a balance between rapid adaptation and maintaining the model's ability to generalize, addressing the limitations of employing either tuning strategy in isolation.

\subsection{Redesigned Query-Key Matching loss}\label{sec:redesigned_query_key_matching}

In the DualPrompt~\citep{wang2022dualprompt}, the matching loss is formalized as $\mathcal{L}_{\text{match}}\left(\boldsymbol{x}, \boldsymbol{k}_t\right) = \gamma\left(q(\boldsymbol{x}), \boldsymbol{k}_t\right)$, with $\gamma$ acting as a distance metric and $q(\boldsymbol{x})$ as a query function. This design aims to minimize the distance between $\boldsymbol{k}_t$ and the query representation of $\boldsymbol{x}$, thus enhancing the affinity between task-specific keys and inputs from corresponding tasks. Nonetheless, this initial formulation overlooks the inter-task relationships, focusing solely on the proximity to a single task key without considering the influence of other task keys. To address this limitation, we propose an enhancement through normalization of similarity scores using a softmax function to ensure that the model's predictions are influenced not only by the nearest task key but also by the relative similarity to all task keys. The revised matching loss incorporating softmax normalization is given by:
\begin{equation}
\label{eqn:matching_loss}
\mathcal{L}_{\text{match}}\left(\boldsymbol{x}, \boldsymbol{k}_t\right) = -\log \left( \frac{e^{\frac{-\gamma\left(q(\boldsymbol{x}), \boldsymbol{k}_t\right)}{\tau}}}{\sum_{i=1}^{t} e^{\frac{-\gamma\left(q(\boldsymbol{x}), \boldsymbol{k}_i\right)}{\tau}}} \right).
\end{equation}
In this updated formula, the temperature factor $\tau$ is introduced to adjust the model's sensitivity to variations in the distance metric, thereby refining the perception of distances between query and key. Furthermore, the modification includes a normalization component in its denominator, aggregating the probabilities that the input $\boldsymbol{x}$ aligns with each of the task-specific keys $\boldsymbol {k}_i$. This enhancement facilitates a more balanced and thorough evaluation of the input's affiliation with all tasks, optimizing the alignment between task-specific inputs and their corresponding keys and addressing the initial formulation's limitation by considering the relative similarity to all task keys.

Similar to the DualPrompt approach, for evaluating a test example \( x \), we select the most appropriate task key index by identifying the minimum distance between \( q(\boldsymbol{x}) \) and \( \boldsymbol{k}_t \), as determined by the criterion \(\text{argmin}_t \gamma(q(\boldsymbol{x}), \boldsymbol{k}_t)\).

\subsection{Training Objective}\label{sec:training_objectives}
The comprehensive training and testing processes are outlined in Algorithm~\ref{alg:train} and Algorithm~\ref{alg:test}, respectively. The objective for the two-stage decoupled training is formulated as follows:
\begin{equation}
\begin{aligned}
&\text{Stage 1:}\quad \min _{\boldsymbol{p}^S, \boldsymbol{p}^T, \phi} \mathcal{L}\left(f_{\phi} \text{{\scriptsize }}\left(f_{\boldsymbol{p}^T,\boldsymbol{p}^S, \boldsymbol{\theta}_T,\boldsymbol{\theta}_S}(\boldsymbol{x})\right), y\right), \\
&\text{Stage 2:}\quad \min _{\theta_T, \theta_S, \boldsymbol{k}_t, \phi} \mathcal{L}\left(f_{\phi} \text{{\scriptsize }}\left(f_{\boldsymbol{p}^T,\boldsymbol{p}^S, \boldsymbol{\theta}_T,\boldsymbol{\theta}_S}(\boldsymbol{x})\right), y\right)+\lambda \mathcal{L}_{\text {match}}\left(\boldsymbol{x}, \boldsymbol{k}_t\right), \label{eqn:loss}
\end{aligned}
\end{equation}
where $f_{\phi}$ denotes the classification head parametrized by $\phi$, $f_{\boldsymbol{p}^T, \boldsymbol{p}^S, \boldsymbol{\theta}_T, \boldsymbol{\theta}_S}(\boldsymbol{x})$ represents the forward process of the adapted, frozen pretrained vision model. It integrates spatial-temporal adapters and prompts with parameters $\boldsymbol{\theta}_T, \boldsymbol{\theta}_S$ and $\boldsymbol{p}^S, \boldsymbol{p}^T$, respectively, as described in Equation~\ref{eqn:forward}. $\mathcal{L}$ is the cross-entropy loss, $\mathcal{L}_{\text{match}}$ is the matching loss defined in Equation~\ref{eqn:matching_loss}, and $\lambda$ is a scalar balancing factor.

\begin{algorithm}[ht]
\caption{DPAT at Training Stage with Decoupled Prompt and Adapter Optimization}
\label{alg:train}
\begin{algorithmic}[1]
    \REQUIRE {Pre-trained ViT backbone \(f\), classification head \(f_\phi\), number of tasks \(N\), training set \(\boldsymbol{D}=\{D_t\}_{t=1}^T\).}
    \REQUIRE {Temporal prompt  \(\boldsymbol{p}^T=\{\boldsymbol{g}^T,\boldsymbol{e}^T\}\) that contains task-agnostic prompt \(g^T\) and task-specific prompt \(\left\{\boldsymbol{e}_t^T\right\}_{t=1}^N\),\\\qquad\quad spatial prompt \(\boldsymbol{p}^S=\{\boldsymbol{g}^S,\boldsymbol{e}^S\}\), a pool of task keys \(\mathbf{K}=\left\{\boldsymbol{k}_t\right\}_{t=1}^N\).}
    \REQUIRE Adapter-S parameterized with \(\theta_S\), Adapter-T parameterized with \(\theta_T\).

    \REQUIRE Initialize $\phi, \boldsymbol{p}^T, \boldsymbol{p}^S, \boldsymbol{\theta}_T, \boldsymbol{\theta}_S, \mathbf{K}$.
    \FOR{$t=1$ \TO $N$} 
        \STATE Select the task-specific prompt $\boldsymbol{e}_t = \{\boldsymbol{e}_t^T, \boldsymbol{e}_t^S\}$ and corresponding task key $\boldsymbol{k}_t$.
        \STATE {Generate the prompted architecture $f_{\boldsymbol{p}^T,\boldsymbol{p}^S, \boldsymbol{\theta}_T,\boldsymbol{\theta}_S}$: attach temporal prompt $\boldsymbol{p}^T =\{\boldsymbol{g}^T,\boldsymbol{e}_t^T\}$ and spatial $\boldsymbol{p}^S=\{\boldsymbol{g}^S,\boldsymbol{e}_t^S\}$ at specified locations within the backbone model $f$, adapting the structure to incorporate spatial and temporal prompts effectively.}

        \FOR{$e=1$ \TO $M_t$} 
            \STATE Draw a mini-batch $B=\{(v_t^b, y_t^b)\}_{b=1}^{|B|}$ from $D_t$.
            \FORALL{$(\boldsymbol{x}, y)$ in $B$}
                \STATE Calculate the adapted feature by $f_{\boldsymbol{p}^T,\boldsymbol{p}^S, \boldsymbol{\theta}_T,\boldsymbol{\theta}_S}(x)$  \hfill $\triangleright$         Stage 1 Optimization
                \STATE Calculate the per sample loss $\mathcal{L_x}$ via Equation.~\ref{eqn:loss}
            \ENDFOR
            \STATE Update $\phi, \boldsymbol{p}^S, \boldsymbol{p}^T$ by backpropagation.
        \ENDFOR
        
        \FOR{$e=1$ \TO $M_t$}
            \STATE Draw a mini-batch $B=\{(v_t^b, y_t^b)\}_{b=1}^{|B|}$ from $D_t$
            \FORALL{$(\boldsymbol{x}, y)$ in $B$}
                \STATE Calculate the adapted feature by $f_{\boldsymbol{p}^T,\boldsymbol{p}^S, \boldsymbol{\theta}_T,\boldsymbol{\theta}_S}(\boldsymbol{x})$ \hfill $\triangleright$         Stage 2 Optimization
                \STATE Calculate the per sample loss  via Equation.~\ref{eqn:loss}
            \ENDFOR
            \STATE Update $\boldsymbol{\theta}_T, \boldsymbol{\theta}_S, \boldsymbol{k}_t, \phi$ by backpropagation.
        \ENDFOR
    \ENDFOR
\end{algorithmic}
\end{algorithm}

\begin{algorithm}
\caption{DPAT at Testing Stage}
\label{alg:test}
\begin{algorithmic}[1]
\REQUIRE Pre-trained ViT backbone \(f\), classification head \(f_\phi\)
    \REQUIRE Temporal prompt  \(\boldsymbol{p}^T=\{\boldsymbol{g}^T,\boldsymbol{e}^T\}\), spatial prompt \(\boldsymbol{p}^S=\{\boldsymbol{g}^S,\boldsymbol{e}^S\}\), a pool of task keys \(\mathbf{K}=\left\{\boldsymbol{k}_t\right\}_{t=1}^N\).
    \REQUIRE Adapter-S parameterized with \(\theta_S\), Adapter-T parameterized with \(\theta_T\).
\REQUIRE Test sample \(\boldsymbol{x}\)
\STATE Generate query feature \(q(\boldsymbol{x})\)
\STATE \(t_x = \arg\min_t \gamma(q(\boldsymbol{x}), \boldsymbol{k}_t)\) \hfill $\triangleright$ Matching for the index of task-specific prompt
\STATE Select the task-specific Prompt \(\boldsymbol{e}_{t_x} = \{\boldsymbol{e}_{t_x}^T,\boldsymbol{e}_{t_x}^S\}\)
\STATE Generate the prompted architecture $f_{\boldsymbol{p}^T,\boldsymbol{p}^S, \boldsymbol{\theta}_T,\boldsymbol{\theta}_S}$: attach temporal prompt $\boldsymbol{p}^T =\{\boldsymbol{g}^T,\boldsymbol{e}_{t_x}^T\}$ and spatial $\boldsymbol{p}^S=\{\boldsymbol{g}^S,\boldsymbol{e}_{t_x}^S\}$ at specified locations within the backbone model $f$, adapting the structure to incorporate spatial and temporal prompts effectively.
\STATE Predict with \(f_{\phi} \text{{\scriptsize }}\left(f_{\boldsymbol{p}^T,\boldsymbol{p}^S, \boldsymbol{\theta}_T,\boldsymbol{\theta}_S}(\boldsymbol{x})\right)\)
\end{algorithmic}
\end{algorithm}

\section{Experiments}

In this section, we first study the performance of DPAT compared to the baselines. We then examine the significance of each component of DPAT through ablation studies.

\subsection{Experiments Settings}
 
\textbf{Datasets.} We evaluate our method across three public datasets: Kinetics-400~\citep{kay2017kinetics} and ActivityNet~\citep{caba2015activitynet} for standard action recognition, alongside EPIC-Kitchens-100~\citep{Damen2021PAMI} for fine-grained action recognition. Following a strategy akin to the vCLIMB \citep{villa2022vclimb} benchmark guidelines, we organize the data into a class-incremental setting. In this arrangement, each dataset's classes are introduced sequentially across a series of 10 tasks, with careful measures taken to prevent class overlap. This structure rigorously evaluates our method's adaptability and learning evolution in a systematic, incremental manner. Further details are provided in Appendix~\ref{append:data}.

\textbf{Evluation Metrics.} We use the widely recognized evaluation metrics of average accuracy (Acc) and backward forgetting (BWF) to measure the model's overall performance across tasks and the extent to which it retains knowledge from previous tasks after learning new ones, respectively, as defined below:
\begin{align}
\centering
&\text{Acc} = \frac{1}{N} \sum_{i=1}^{N} R_{N, i}, &&\text{BWF} = \frac{1}{N-1} \sum_{i=1}^{N-1} (R_{i, i} - R_{N, i}),
\end{align}
 where $N$ represents the total number of tasks, $R_{N, i}$ denotes the accuracy of the model on task $i$ after learning all $N$ tasks, and $R_{i, i}$ represents the accuracy of the model on task $i$ immediately after learning it.

\textbf{Implementation.} We leverage the ViT-B/16 architecture~\citep{dosovitskiy2020image}, pre-trained on the ImageNet-21K (IN-21K) dataset, as the backbone for DPAT. To adapt to task-specific requirements while maintaining the integrity of the pre-trained features, we freeze the backbone and sequentially fine-tune adapters and prompts. Optimization is performed using the Adam~\citep{kingma2014adam} optimizer, with a batch size of 64 across 50 epochs for each task. We set the bottleneck ratios for spatial and temporal adapters at 0.25 and the scaling factor, $\lambda$, to 1. To ensure robustness and generalizability of our findings, we select benchmark methods that operate under analogous conditions, facilitating a rigorous and fair comparison.

\subsection{Comparison with baseline}
The comparative analysis incorporates methodologies such as EWC~\citep{kirkpatrick2017overcoming}, MAS~\citep{aljundi2018memory}, iCaRL~\citep{rebuffi2017icarl}, and the state-of-the-art PIVOT~\citep{villa2023pivot}. To ensure a fair comparison, configurations for EWC, MAS, and iCaRL are enhanced with a tunable adapter. In conducting a comparison with PIVOT, our analysis adopts the CLIP~\citep{radford2021learning} ViT-B/16 model as the underlying architecture, aligning with the image encoder utilized by PIVOT itself. 

\textbf{Results on Kinetics-400 and ActivityNet.} As illustrated in Table \ref{tab:basic}, DPAT (IN-21K), employing a ViT-B/16 model pre-trained on ImageNet-21K, substantially surpasses traditional rehearsal-free methodologies MAS and EWC. Further analysis reveals that against the state-of-the-art PIVOT, DPAT (CLIP), which also utilizes the same CLIP ViT-B/16 backbone as PIVOT, secures higher prediction accuracy and demonstrates reduced backward forgetting on both Kinetics-400 and ActivityNet datasets, all achieved without relying on replayed video instances. This enhanced performance not only highlights DPAT's adeptness at preserving previously acquired knowledge amidst new data assimilation but also signifies its profound impact on advancing continual learning paradigms.

\textbf{Results on EPIC-Kitchens-100.} Table~\ref{tab:epic} showcases the comparative results on the EPIC-Kitchens-100 dataset. DPAT notably excels in verb prediction against baseline models, underscoring the significant capability of the temporal adapter design to capture temporal information effectively. Furthermore, the reduced backward forgetting observed across models highlights how our innovative prompt design, incorporating a key-query matching strategy, effectively mitigates forgetting in temporal modeling contexts. Although the prediction accuracy for nouns is slightly lower than that of PIVOT, our model still demonstrates the best overall performance in action prediction. This nuanced balance between temporal and spatial knowledge preservation, despite the slight trade-off, underscores DPAT's robustness and adaptability, making it a competitive choice for continual activity recognition in dynamic environments such as those presented by the EPIC-Kitchens-100 dataset.

\begin{table}[ht]
\centering
\caption{Comparative Evaluation Metrics for Models on Kinetics-400 and ActivityNet Datasets: Memory Usage (Mem. (RI), the number of Replayed Instances), Accuracy (Acc $\uparrow$, higher is better), and Backward Forgetting (BWF $\downarrow$, lower is better). DPAT outperforms existing rehearsal-free methods EWC and MAS, replay-based iCaRL, and the PIVOT when employing the same CLIP ViT-B/16 backbone in terms of accuracy and backward forgetting.}
\label{tab:basic}
{\small
\begin{tabular}{@{}ccccccc@{}}
\toprule
\multirow{2}{*}{\textbf{Model}} & \multicolumn{3}{c}{\textbf{Kinetics-400}} & \multicolumn{3}{c}{\textbf{ActivityNet}} \\
\cmidrule(lr){2-4}\cmidrule(l){5-7}
 & Mem. (RI) & Acc $\uparrow$ & BWF $\downarrow$ & Mem. (RI) & Acc $\uparrow$ & BWF $\downarrow$ \\
\cmidrule(lr){2-4}\cmidrule(l){5-7}
iCaRL & 8000 & 48.7\% & 30.3\% & 4000 & 60.9\% & 15.72\%\\
PIVOT & 4000 & 56.1\% & 25.7\% & 2000 & 73.6\% & 11.1\% \\
\cmidrule(lr){2-4}\cmidrule(l){5-7}
EWC & 0 & 25.2\% & 15.1\% & 0 & 11.3\% & 13.1\% \\
MAS & 0 & 23.8\% & 11.6\% & 0 & 10.4\% & 5.5\% \\
DPAT (IN-21K) & 0 & 58.5\% & 24.7\% & 0 & 71.9\% & 12.6\% \\
\textbf{DPAT (CLIP)} & 0 & \textbf{61.3\%} &22.3\% & 0 & \textbf{74.5\%} & 11.2\% \\
\midrule
Upper Bound & 0 & 84.1\% & - & 0 & 89.3\% & - \\
\bottomrule 
\end{tabular}
}
\end{table}

\begin{table}[ht]
\centering
\caption{Performance Analysis on the EPIC-Kitchens-100 Dataset: Accuracy (Acc) and Backward Forgetting (BWF) Metrics for Verb, Noun, and Action Predictions, where an action is deemed correctly predicted only if both the verb and noun components are accurately identified.}
\label{tab:epic}
{\small
\begin{tabular}{cccccccc}
\toprule
\multirow{2}{*}{\textbf{Model}} & \multirow{2}{*}{\textbf{Mem. (RI)}} & \multicolumn{2}{c}{\textbf{Verb}} & \multicolumn{2}{c}{\textbf{Noun}} & \multicolumn{2}{c}{\textbf{Act}} \\
\cmidrule(lr){3-4}\cmidrule(lr){5-6}\cmidrule(l){7-8}
 & & Acc $\uparrow$ & BWF $\downarrow$ & Acc $\uparrow$ & BWF $\downarrow$ & Acc $\uparrow$ & BWF $\downarrow$ \\
\cmidrule(lr){2-2}\cmidrule(lr){3-4}\cmidrule(lr){5-6}\cmidrule(l){7-8}
iCaRL & 5940 & 39.7\% & 25.2\%& 30.8\% & 19.6\%& 17.9\% & 16.4\% \\
PIVOT & 2970 & 46.1\% & 20.2\%& \textbf{43.5}\% & 16.8\%& 29.8\% & 13.2\% \\
\cmidrule(lr){2-2}\cmidrule(lr){3-4}\cmidrule(lr){5-6}\cmidrule(l){7-8}
EWC & 0 & 18.1\% & 10.1\% & 11.1\% & 8.3\% & 5.6\% & 1.1\%\\
MAS & 0 & 19.6\% & 9.3\% & 13.3\% & 7.7\% & 6.7\% & 0.9\%\\
DPAT (IN-21K) & 0 & 51.1\% & 19.1\% & 38.1\% & 17.7\% & 30.3\% & 12.6\%\\
\textbf{DPAT (CLIP)} & 0 & \textbf{53.8\%} & 18.6\% & 42.3\% & 17.1\% & \textbf{32.3}\% & 11.5\%\\ 
\midrule
Upper Bound & 0 & 65.3\% & - & 56.1\% & - &{42.8\%} & {-} \\
\bottomrule
\end{tabular}
}
\end{table}

\subsection{Ablation Studies}
In this section, we conduct ablation studies to scrutinize the characteristics and efficacy of our fundamental design elements.

\textbf{Effect of Model Component.} In the ablation studies summarized in Table~\ref{tab:model components}, we systematically assess the contribution of each component to our model's proficiency in continual action recognition, utilizing the Kinetics-400 dataset for our experiments. The absence of the Temporal Adapter leads to a substantial decrease in accuracy by \(38.2\%\), underscoring its critical role in capturing temporal dynamics crucial for precise action recognition. The complete removal of all adapters results in a marked decrease in accuracy to 25\%, indicating a further decline in learning performance. This underscores the critical role of adapters in adapting to new tasks. Interestingly, this configuration leads to minimal forgetting, suggesting that a combination of prompt learning with a frozen pre-trained model provides a stable foundation, even as it points to the necessity of adapters for effective task adaptation. In contrast, the exclusion of the Prefix Prompt detrimentally impacts accuracy and leads to the highest BWF rate among the configurations tested, emphasizing the prompt's vital importance in bolstering the model's learning stability and its ability to mitigate forgetting efficiently. Derived from experiments on the Kinetics-400 dataset, these findings highlight the intricate balance between leveraging adapters for generalization and prompt tuning for stability. This balance underscores the necessity of designing both components within our model to navigate the challenges of continual learning effectively.

\begin{table}[ht]
    \centering
    \caption{Results of ablation studies on model components}
    \label{tab:model components}
    \begin{tabular}{lcc}
    \toprule
        \textbf{Training Method} & \textbf {Acc} $\uparrow$ & \textbf{BWF} $\downarrow$ \\
    \midrule
        \textbf{DPAT} & \textbf{61.3}\% & 22.3\% \\
        Ablate Temporal Adapter & 33.1\% & 19.5\%\\
        Ablate all Adapter & 25.0\% & 5.1\%\\
        Ablate Task-agnostic Prefix & 58.7\% & 24.1\%\\
        Ablate {All} Prefix Prompt & 48.2\% & 33.6\%\\
   \bottomrule
    \end{tabular}
\end{table}

\begin{figure}[ht]
    \centering
    \includegraphics[width=0.4\textwidth]{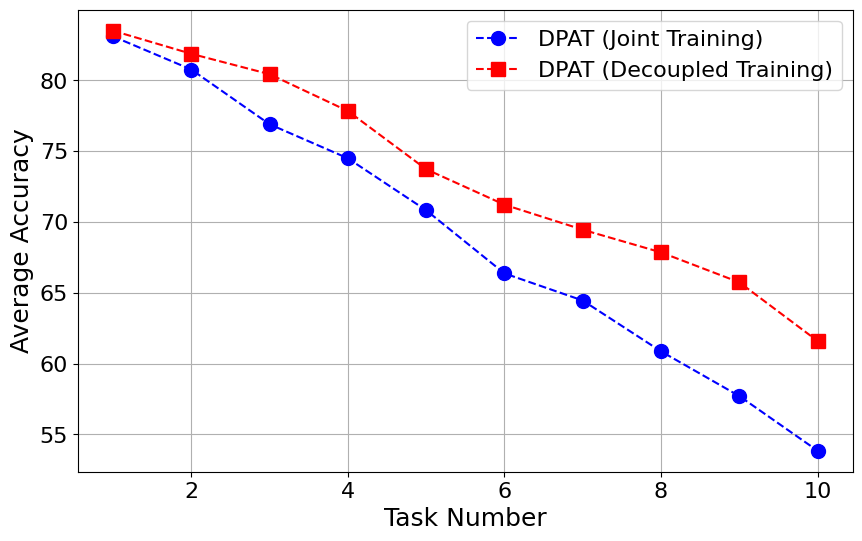}  
    \caption{Comparative Result of DPAT with Joint and Decoupled Training Strategies on Kinetics-400}
    \label{fig:forgetting_curves}
\end{figure}

\textbf{Effect of Decoupled Training.} 
To illustrate the efficacy of our decoupled training strategy, we present the learning curve comparison between decoupled and joint training in Figure~\ref{fig:forgetting_curves}. The x-axis represents the current task, while the y-axis measures the average accuracy (Acc). Conducting experiments on the Kinetics-400 dataset, our findings reveal that although both strategies commence with comparable accuracy levels, the decoupled training demonstrates a trend of slower degradation in mean accuracy alongside reduced backward forgetting. This highlights our model's enhanced capability to preserve previously learned knowledge over time.

\textbf{Effect of Query Matching Loss.} Table~\ref{tab:Query} highlights DPAT's superiority over the DualPrompt model, with DPAT achieving a Matching Accuracy of $45.6\%$ compared to DualPrompt's $32.8\%$ on Kinetics-400. This improvement underscores our optimized matching loss's efficacy in enhancing task-specific contrast, allowing for more precise task-specific key embeddings. Notably, the elevation in Matching Accuracy corresponds with a significant boost in Prediction Accuracy  and a reduction in Backward Forgetting, showcasing the direct impact of improved alignment on model performance and memory retention.

\begin{table}[ht]
\centering
\caption{Effect of Different Query Matching Loss}
\label{tab:Query}
\begin{tabular}{lccc}
\toprule
\textbf{Query Loss} & \textbf{Matching Acc} & \textbf{Acc} $\uparrow$ & \textbf{BWF} $\downarrow$ \\
\midrule
DualPrompt & 32.8\% & 57.2\% & 25.5\% \\
\textbf{DPAT} & \textbf{45.6}\% & 61.3\% &22.3\% \\
\bottomrule
\end{tabular}
\end{table}

\section{Conclusion}
In this paper, we present a simple yet effective rehearsal-free approach for continual activity recognition. Eliminating the necessity for integrating auxiliary temporal architectures, relying on external modal inputs, or undertaking extensive fine-tuning, our architecture employs adapters in conjunction with prompt tuning to excel in the realm of continual action recognition, leveraging a frozen pre-trained model. Furthermore, We introduce a decoupled training strategy that capitalizes on the adapter's generalization capabilities and the stability provided by prompt tuning, effectively mitigating the issue of forgetting. Evaluations on various challenging benchmarks for continual action recognition indicate that our model performs well in comparison to existing methods. 

Despite the considerable advantages offered by our methodology, it is not without its limitations. The reliance of our task-specific prompt design on predefined task boundaries precludes its application in an online learning context. Our approach currently operates within a closed-set classification paradigm, transitioning to a more realistic open-set continual action recognition setting stands as a promising avenue for future research.

\subsubsection*{Acknowledgments}
This research is supported by an Academic Research Grant
No. MOE-T2EP20121-0015 from the Ministry of Education
in Singapore.

\bibliography{collas2024_conference}
\bibliographystyle{collas2024_conference}


\appendix
\section{Appendix}

\subsection{Datasets} \label{append:data}
\textbf{Kinetics-400.} Kinetics-400 stands as an extensive dataset tailored for action recognition, encompassing roughly 300k video snippets classified into 400 distinct human action categories. Originating from a wide variety of YouTube videos, each clip is meticulously trimmed to approximately 10 seconds to ensure uniformity. For the purposes of this work, the dataset is partitioned into 10 separate tasks, with each task consists of 40 unique action classes, arranged in a sequential, class-incremental manner.

\textbf{ActivityNet.} 
ActivityNet is a large-scale dataset designed for action recognition, featuring over 20,000 video clips spread across 200 activity classes. Sourced from YouTube, it offers a wide-ranging and diverse collection of real-world scenarios that capture a broad spectrum of everyday human activities. Similar to Kinetics-400, we organize the dataset into a class-incremental setting with 10 tasks, ensuring a neat division of 20 activities per task. 

\textbf{Epic-Kitchen-100. }  Epic-Kitchen-100 is a large-scale egocentric video dataset that records over 100 hours of kitchen unscripted activities. It consists of 90K action segments, which are split into train/val/test sets of 67K/10K/13K. Differ from preceding two datasets, Epic-Kitchen-100 defines an action as a combination of a verb and a noun. Given the necessity to match both verbs and nouns for action recognition, this task presents a higher level of complexity compared to action recognition in previous datasets characterized by single label prediction. We employ a class-incremental strategy for verb classification, dividing 97 verb categories into 10 non-overlapping tasks to systematically introduce new classes. Concurrently, noun prediction is approached with a task-incremental strategy, wherein all tasks share a consistent set of total 300 noun classes. This deliberate division, prioritizing verbs for the class-incremental learning setting, stems from our intent to scrutinize the model's ability to discern and learn from the nuanced temporal dynamics across different tasks.

\subsection{Implementation Details}
Following the ViT architecture guidelines, we sample videos to 16 frames. Each frame is then randomly cropped and resized to $224 \times 224$ pixels. Additionally, we enhance the diversity of the training data by applying data augmentation techniques, including mixup~\citep{zhang2017mixup}, label smoothing, horizontal flipping, color jittering, and RandAugment~\citep{cubuk2020randaugment}. {Specifically, for mixup, an alpha of 0.2 was utilized. Label smoothing was implemented using a factor of 0.1. In the color jittering process, brightness, contrast, and saturation adjustments were uniformly set to 0.4, with hue adjustments at 0.1. We deployed RandAugment with a configuration of 2 transformations at a magnitude level of 10.} Finally, to ensure consistency across diverse datasets, we normalize each frame to the range $[0, 1]$

As depicted in Figure~\ref{fig:sub1}, our method, DPAT, integrates spatial and temporal adapters into every ViT block. {The architecture of the adapter adheres to a bottleneck design, incorporating two fully connected (FC) layers separated by an activation layer. The design involves diminishing the input dimensionality through the initial FC layer, and subsequently restoring it via the second FC layer. The extent of dimensionality reduction is determined by a bottleneck ratio—defined as the quotient of the bottleneck to the input dimension—with a consistent ratio of 0.25 applied throughout the experiment. }

Task-agnostic prompts, denoted as $\boldsymbol{g}^T$ and $\boldsymbol{g}^S$, are introduced in the first two blocks of the pre-trained ViT model. Conversely, task-specific prompts, represented by $\boldsymbol{e}^T$ and $\boldsymbol{e}^S$, target the third through fifth ViT blocks. {Our analysis revealed an optimal setup consisting of task-specific prompts with a length of 5, alongside task-agnostic prompts extended to a length of 20.}

DPAT employs a sequential optimization strategy, optimizing adapters and prompts using the Adam optimizer in two separate stages. We utilize a batch size of 64 and a total of 50 epochs for both training stage, starting with a base learning rate of $1 \times 10^{-3}$ for prompt tuning and $3 \times 10^{-4}$ for adapter tuning, followed by a cosine decay schedule. Moreover, the balancing scaling factor $\lambda$ was selected from the set \{0.01, 0.1, 1, 10\}. Following a similar parameter search for the temperature scaling factor $\tau$ among \{0.01, 0.05, 0.1, 0.5\}, we determined that $\lambda = 1$ and $\tau = 0.1$ provide the optimal outcomes.

\subsection{Additional Experiments and Analysis}
\label{sec:appendixA}
In this section, we present additional experiments and ablation studies to further validate the effectiveness of our proposed DPAT approach and investigate the impact of various components and hyperparameters on its performance.

\subsubsection{Ablation Studies on ActivityNet}
\label{subsec:A3}

To further validate the effectiveness of our proposed DPAT approach and the individual contributions of its components, we conducted additional ablation studies on the ActivityNet dataset. The results are summarized in Table~\ref{tab:ablationActivityNet}, highlighting the importance of both the temporal adapter and the task-agnostic prefixes within our framework.

\begin{table}[htbp]
\centering
\caption{Ablation studies on the ActivityNet dataset.}
\label{tab:ablationActivityNet}
\begin{tabular}{lcc}
\toprule
\textbf{Method} & \textbf{Acc$\uparrow$} & \textbf{BWF $\downarrow$} \\
\midrule
\textbf{DPAT} & \textbf{74.5} & 11.2 \\
Ablate Temporal Adapter & 41.2 & 15.7 \\
Ablate All Adapters & 32.6 & \textbf{6.8} \\
Ablate Task-agnostic Prefix & 68.1 & 13.3 \\
Ablate All Prefixes & 59.4 & 19.5 \\
\bottomrule
\end{tabular}
\end{table}

The findings on the ActivityNet dataset corroborate those observed on the Kinetics-400 dataset, underscoring the critical role that both adapters and prefixes play in our methodology. The removal of the temporal adapter or all adapters results in a significant decline in accuracy, whereas the elimination of task-agnostic prefixes or all prefixes exacerbates backward forgetting and diminishes overall performance.

\subsubsection{Experiments on UCF-101}
\label{subsec:UCF101}

Further experiments were conducted on the UCF-101 dataset to offer a broader evaluation of our approach. The outcomes, detailed in Table~\ref{tab:UCF101}, show our method's performance in comparison to existing benchmarks.

\begin{table}[htbp]
\centering
\caption{Performance comparison on the UCF-101 dataset.}
\label{tab:UCF101}
\begin{tabular}{lccc}
\toprule
\textbf{Model} & \textbf{Mem. (RI)} & \textbf{Acc} $\uparrow$ & \textbf{BWF} $\downarrow$  \\
\midrule
iCaRL & 2020 & 82.1 & 12.3 \\
\textbf{PIVOT} & 1010 & \textbf{94.1} & \textbf{3.7} \\
EWC & 0 & 15.2 & 30.1 \\
MAS & 0 & 16.8 & 8.6 \\
DPAT (IN-21K) & 0 & 89.2 & 4.3 \\
DPAT (CLIP) & 0 & 92.8 & 3.9 \\
\midrule
Upperbound & -- & 96.8 & -- \\
\bottomrule
\end{tabular}
\end{table}

While our DPAT approach did not outperform PIVOT on the UCF-101 dataset, the results were closely matched. The superior performance of PIVOT might be attributed to its use of a replay buffer, which plays a pivotal role in mitigating forgetting and enhancing model performance on this relatively simpler dataset. Notably, our method requires no additional memory for data storage, presenting an advantage in complex continual learning scenarios for which it was designed.

\subsubsection{Ablation Studies on Prompt Position}
\label{subsec:PromptPosition}

A parameter search was conducted to fine-tune the positioning of temporal and spatial prompts within the DPAT framework, with the starting position ($\text{start}_g$) set to 1 and the ending position ($\text{end}_e$) set to 5. We varied the ending position ($\text{end}_g$) and adjusted the starting position of the subsequent element ($\text{start}_e$) accordingly. The search results, provided in Table~\ref{table:prompt_position}, validate our initial configuration choices, though it is acknowledged that due to the expansive search space, these findings may not represent the optimal configuration.

\begin{table}[htbp]
\centering
\caption{Ablation studies on prompt position.}
\label{table:prompt_position}
\begin{tabular}{cc}
\toprule
$\textbf{end}_g$ & \textbf{Acc} \\
\midrule
1 & 59.2 \\
\textbf{2} & \textbf{61.3} \\
3 & 60.9 \\
4 & 58.8 \\
\bottomrule
\end{tabular}
\end{table}

\subsubsection{Ablation studies on  Bottleneck Ratio}
\label{subsec:OptimalBottleneckRatio}

The bottleneck ratio, a critical hyperparameter in our DPAT framework, influences the trade-off between model capacity and efficiency. Experiments were conducted on the Kinetics-400 dataset to identify the optimal bottleneck ratio. Various ratios were tested, and their impacts on average accuracy and backward forgetting were assessed. As Table~\ref{table:bottleneck_ratio} indicates, a bottleneck ratio of 0.25 optimally balances model capacity with efficiency, leading to the best performance on this dataset.

\begin{table}[htbp]
\centering
\caption{Ablation studies on bottleneck ratio on Kinetics-400.}
\label{table:bottleneck_ratio}
\begin{tabular}{ccc}
\toprule
\textbf{Bottleneck Ratio} & \textbf{Acc} $\uparrow$  & \textbf{BWF} $\downarrow$  \\
\midrule
0.05 & 59.7 & 21.8 \\
0.1 & 60.5 & 22.1 \\
\textbf{0.25} & \textbf{61.3} & 22.3 \\
0.5 & 60.9 & 22.7 \\
\bottomrule
\end{tabular}
\end{table}

\end{document}